\documentclass[a4paper,10pt]{article}

\usepackage{amsmath}  % newly added package
\usepackage{verbatim} % newly added package
\usepackage{amsfonts}
\usepackage{amssymb}
\usepackage{graphicx}
\usepackage{booktabs}
\usepackage{paralist}
\usepackage{url}
\newcommand{\urlBiBTeX}[1]{\url{#1}}

\DeclareMathOperator{\val}{=}  % for p=v atoms
\def\happens{\textsf{\small happensAt}}
\def\happensFor{\textsf{\small happensFor}}
\def\initially{\textsf{\small initially}}
\def\holdsAt{\textsf{\small holdsAt}}
\def\holdsFor{\textsf{\small holdsFor}}
\def\initiatedAt{\textsf{\small initiatedAt}}
\def\terminatedAt{\textsf{\small terminatedAt}}

\def\terminates{\textsf{\small terminates}}
\def\broken{\textsf{\small broken}}

\def\intersectall{\textsf{\small intersect\_all}}
\def\complement{\textsf{\small complement}}

\def\nbf{\textsf{\small not}}
\def\true{\textsf{\small true}}
\def\false{\textsf{\small false}}

\newenvironment{mysplit}%
  {\arraycolsep 0pt \begin{array}{l}}%
  {\end{array}}

\def\immobile{$\mathit{immobile}$}
\def\meet{$\mathit{meeting}$}
\def\fight{$\mathit{fighting}$}
\def\leave{$\mathit{leaving\_object}$}
\def\move{$\mathit{moving}$}

\title{A Logic Programming Approach to Activity Recognition}

\author{Alexander Artikis$^1$, Marek Sergot$^2$ and Georgios Paliouras$^1$ \\
       $^1$Institute of Informatics \& Telecommunications, \\
       NCSR Demokritos, Athens 15310, Greece\\
       $^2$Department of Computing, Imperial College London, SW7 2BT, UK\\
       \texttt{\{a.artikis, paliourg\}@iit.demokritos.gr, mjs@doc.ic.ac.uk}
}
\date{}
\begin{document}
\maketitle

\begin{abstract}
We have been developing a system for recognising human activity given a symbolic representation of video content. The input of our system is a set of time-stamped short-term activities detected on video frames. The output of our system is a set of recognised long-term activities, which are pre-defined temporal combinations of short-term activities. The constraints on the short-term activities that, if satisfied, lead to the recognition of a long-term activity, are expressed using a dialect of the Event Calculus. We illustrate the expressiveness of the dialect by showing the representation of several typical complex activities. Furthermore, we present a detailed evaluation of the system through experimentation on a benchmark dataset of surveillance videos.
\end{abstract}

\section{Introduction}
%\vspace{-.1cm}

A common approach to human activity recognition separates low-level from high-level recognition. The output of the former type of recognition is a set of activities taking place in a short period of time: `short-term activities'. The output of the latter type of recognition is a set of  `long-term activities', ie pre-defined temporal combinations of short-term activities. We focus on high-level recognition.

We define a set of long-term activities of interest, such as `fighting' and `meeting', as temporal combinations of short-term activities --- eg, `walking', `running', and `inactive' (standing still) --- using a logic programming implementation of the Event Calculus \cite{kowalski86}. More precisely, we employ the Event Calculus to express the temporal constraints on a set of short-term activities that, if satisfied, lead to the recognition of a long-term activity.

We presented preliminary results on activity recognition from video content in \cite{artikis09AIAIshort}. (In  \cite{artikisIJAITshort} we described some initial steps towards automatically constructing activity definitions using machine learning techniques --- the use of such techniques is out of the scope of this paper.) In this paper we extend our previous work in the following ways.
First, we use a more efficient Event Calculus dialect and implementation to compute the intervals of long-term activities. 
%This dialect outperforms, for retrospective recognition, the Cached Event Calculus \cite{chittaro96} which has been specifically developed for efficient temporal reasoning. We also outline how the proposed Event Calculus dialect may be extended in order to be used for on-line recognition.
%
Second, we illustrate the expressiveness of the proposed Event Calculus dialect by presenting several complex activity definitions. We are able to  construct much more succinct representations of activity definitions for video surveillance than we had in our earlier work.
Third, we present a more detailed and informative evaluation of the Event Calculus on activity recognition. We show through experimentation how incomplete short-term activity narratives, inconsistent annotation of short-term and long-term activities, and a limited dictionary of short-term activities and context variables affect recognition accuracy.
Fourth, we evaluate our approach on a dataset with a refined dictionary of short-term activities, in order to validate experimentally our intuition that a finer classification of short-term activities increases, under certain circumstances, the accuracy of long-term activity recognition. Indeed, the refined dictionary of short-term activities --- which can be provided by state-of-the-art short-term activity recognition systems --- together with the updated long-term activity definitions presented in this paper, lead to much higher Precision and Recall rates.
%
%Fifth, we present a considerably more detailed and extensive comparison of our approach to related work. %We outline the benefits of our approach with respect to purely temporal reasoning systems as well as logic programming approaches specifically developed for activity recognition from video content.

The remainder of the paper is organised as follows. First, we present the Event Calculus dialect that we employ to formalise activity definitions. Second, we describe the dataset of short-term activities on which we perform long-term activity recognition. Third, we present our knowledge base of long-term activity definitions. Fourth, we present our experimental results. Finally, we discuss related work and outline directions for further research.

\vspace{-.4cm}
\section{The Event Calculus} \label{sec:ec}
\vspace{-.1cm}

Our long-term activity recognition (LTAR) system consists of a logic programming (Prolog) implementation of an Event Calculus dialect. The Event Calculus, introduced by Kowalski and Sergot \cite{kowalski86}, is a many-sorted, first-order predicate calculus for representing and reasoning about events and their effects. For the dialect used here, hereafter LTAR-EC (event calculus for long-term activity recognition), the time model is linear and it may include real numbers or integers. Where $F$ is a \emph{fluent} --- a property that is allowed to have different values at different points in time --- the term $F \val V$ denotes that fluent $F$ has value $V$. Boolean fluents are a special case in which the possible values are \true\ and \false. Informally, $F \val V$ holds at a particular time-point if $F \val V$ has been \emph{initiated} by an event at some earlier time-point, and not \emph{terminated} by another event in the meantime.

%\vspace{-.4cm}

\begin{table}[t]
\caption{Main Predicates of the LTAR-EC.}\label{tbl:ec}\vspace{-.2cm}
\begin{center}
\renewcommand{\arraystretch}{0.9}
\setlength\tabcolsep{3pt}
\begin{tabular}{ll}
\hline\noalign{\smallskip}
\multicolumn{1}{c}{\textbf{Predicate}} & \multicolumn{1}{c}{\textbf{Meaning}}  \\
\noalign{\smallskip}
\hline
\noalign{\smallskip}
$\happens(E,\ T)$ & Event $E$ is occurring at time $T$  \\[2pt]

$\happensFor(E,\ I)$ & $I$ is the list of maximal intervals \\
& during which event $E$ takes place  \\[2pt]

$\initially(F \val V)$ & The value of fluent $F$ is $V$ at time $0$  \\[2pt]

$\holdsAt(F \val V,\ T)$ & The value of fluent $F$ is $V$ at $T$ \\[2pt]

$\holdsFor(F \val V,\ I)$ & $I$ is the list of maximal intervals \\
                           & for which $F\val V$ holds continuously\\[2pt]

$\initiatedAt(F \val V,\ T)$ & At time $T$ a period of time  \\
&  for which $F\val V$ is initiated \\[2pt]

$\terminatedAt(F \val V,\ T)$ & At time $T$ a period of time \\
&  for which $F\val V$ is terminated \\
\hline
\end{tabular}
\end{center}\vspace{-.8cm}
\end{table}

An \emph{event description} in LTAR-EC includes axioms that define, among other things, the event occurrences (with the use of the \happens\ and \happensFor\ predicates), the effects of events (with the use of the \initiatedAt\ and \terminatedAt\ predicates), and the values of the fluents (with the use of the \initially, \holdsAt\ and \holdsFor\ predicates). Table \ref{tbl:ec} summarises the main predicates of LTAR-EC. Variables, starting with an upper-case letter, are assumed to be universally quantified unless otherwise indicated. Predicates, function symbols and constants start with a lower-case letter.

%Almost all published accounts of the Event Calculus focus exclusively on the formulation of the domain-independent axioms for \holdsAt. Computationally, this provides a procedure for evaluating queries of the form $\holdsAt(F \val V, T)$ for determining the value $V$ of a fluent $F$ at some specified time-point $T$. In event recognition (as we have formulated it), and in many other kinds of applications, however, we are interested primarily in the inverse problem, ie in computing the time-points at which a given fluent has a particular value. In the notation of this paper, that is a query of the form $\holdsFor(F \val V, I)$ where $I$ is the list of maximal intervals for which $F$ has value $V$ continuously. This problem is not much discussed in the literature.

The domain-independent axioms for \holdsAt\ and \holdsFor\ are such that, for any fluent $F$, $\holdsAt(F \val V,\ T)$ if and only if time-point $T$ belongs to one of the maximal intervals of $I$ such that $\holdsFor(F \val V,\ I)$. However, for efficiency the implementation employs different procedures for these two tasks, and various indexing techniques to reduce search and improve  efficiency further.  Briefly, to compute $\holdsFor(F \val V,\ I)$, we find all time-points $T_i$ in which $F\val V$ is initiated, and then, for each $T_i$, we compute the first time-point after $T_i$ in which $F\val V$ is terminated. If the list of initiating time-points is generated in sorted order, which is easy to arrange,  both steps can make effective use of indexing. In particular, if the list of initiating time-points contains an adjacent pair $\dots, T_i, T_{i+1}, \dots$ then the terminating time-point corresponding to $T_i$ must occur between $T_i$ and $T_{i+1}$. In outline, the indexing works as follows.

The domain-independent axioms for \holdsAt\ can be written in the following form:
\begin{align}
% &
% \label{eq:ec-holdsAt-init}
% \begin{mysplit}
% \holdsAt(\,F\val V,\ T\,) \leftarrow \\
% \quad   \initially(\,F\val V\,),\\
% \quad   \nbf\ \broken(\,F\val V,\ 0,\ T\,)
% \end{mysplit}\\
&
\label{eq:ec-holdsAt}
\begin{mysplit}
\holdsAt(F\val V,\ T) \leftarrow \\
\quad   \initiatedAt(F\val V,\ T_s),\  \nbf\ \broken(F\val V,\ T_s,\ T)
\end{mysplit}\\
% \intertext{where}
& \label{eq:ec-broken-terminated}
\begin{mysplit}
\broken(F\val V,\, T_s,\, T) \leftarrow \\
\quad   \terminatedAt(F\val V,\ T_f),\ T_s < T_f < T
\end{mysplit}\\
& \label{eq:ec-broken-initiated}
\begin{mysplit}
\broken(F\val V_1,\ T_s,\ T) \leftarrow \\
\quad  \initiatedAt(F\val V_2,\ T_f),\ V_1 \neq V_2, \ T_s < T_f < T
\end{mysplit}
\end{align}
\nbf\ in rule~\eqref{eq:ec-holdsAt} represents `negation by failure', which provides a form of default persistence (`inertia') of fluents.

According to rule \eqref{eq:ec-broken-terminated}, a period of time for which $F\val V$ holds is \textsf{broken} at $T_f$ if $F\val V$ is terminated at  time $T_f$.
According to rule \eqref{eq:ec-broken-initiated}, if $F \val V_2$ is initiated at $T_f$ then effectively $F \val V_1$ is terminated at time $T_f$, for all other possible values $V_1$ of $F$. Rule \eqref{eq:ec-broken-initiated} ensures therefore that a fluent cannot have more than one value at any time.
%(We do not insist that a fluent must have a value at every time-point. In CER-EC there is a difference between initiating a Boolean fluent $F \val \false$ and terminating $F \val \true$: the first implies, but is not implied by, the second.)

Besides the general, domain-independent rule $\initiatedAt(F \val V,\ 0) \leftarrow \initially(F \val V)$, the definitions of \initiatedAt\ and \terminatedAt\ are domain specific. One common form of rule for \initiatedAt, eg,  has the general form:
\begin{align}
& \label{eq:ec-initiatedAt-typical}
\begin{mysplit}
\initiatedAt(F\val V,\ T) \leftarrow \\
\quad  \happens(Ev,\ T), \ \mathit{Conditions}[T]
\end{mysplit}
\end{align}
where $\mathit{Conditions}[T]$ is some set of further conditions referring to time $T$. Concrete examples of \initiatedAt\ rules are provided in the section that follows.

To explain what we mean by indexing, note that
clauses \eqref{eq:ec-holdsAt}, \eqref{eq:ec-broken-terminated} and \eqref{eq:ec-broken-initiated} can be written equivalently as follows:
\begin{align}
& \label{eq:ec-holdsAt-imp}
\begin{mysplit}
\holdsAt(F\val V,\ T) \leftarrow \\
\quad   \initiatedAt(F\val V,\ 0,\ T_s,\, T),\ \nbf\ \broken(F\val V,\ T_s,\ T)
\end{mysplit}\\
& \label{eq:ec-broken-terminated-imp}
\begin{mysplit}
\broken(F\val V,\ T_{min},\ T_{max})  \leftarrow \\
\quad   \terminatedAt(F\val V,\ T_{min},\ T_f,\ T_{max})
\end{mysplit}\\
& \label{eq:ec-broken-initiated-imp}
\begin{mysplit}
\broken(F\val V,\ T_{min},\ T_{max}) \leftarrow \\
\quad  \initiatedAt(F\val V_2,\ T_{min},\ T_f,\ T_{max}),\ V_1 \neq V_2
\end{mysplit}
\intertext{when every rule of the form~\eqref{eq:ec-initiatedAt-typical} is transformed into the form:}
& \label{eq:ec-initiatedAt-typical-imp}
\begin{mysplit}
\initiatedAt(F\val V,\ T_{min},\ T,\, T_{max}) \leftarrow \\
\quad  \happens(Ev,\ T_{min},\ T,\, T_{max}), \  \mathit{Conditions}[T]
\end{mysplit}
\end{align}
The extra arguments in \initiatedAt, \terminatedAt, and \happens\ specify the range of time-points $T_{min}$ and $T_{max}$ between which the time-point $T$ of interest must occur. Thus
$\happens(Ev,$ $ T_{min},\ T,\, T_{max})$ iff $\happens(Ev,\ T)$ and $T_{min} < T < T_{max}$.

Our implementation automatically transforms \initiatedAt\ rules of the form~\eqref{eq:ec-initiatedAt-typical} into the form~\eqref{eq:ec-initiatedAt-typical-imp} on compilation, in a process transparent to the user.
The advantage is that in Prolog execution of $\holdsAt$ and $\holdsFor$, when $\happens(Ev,\ T_{min},\ T,\, T_{max})$ is  called,  $Ev$ is always ground (variable-free), which exploits Prolog's built-in indexing when searching for occurrences of event $Ev$. But much more importantly, $T_{min}$ and $T_{max}$ are also always ground which means that the storage of  \happens\ data can be indexed to exploit this and reduce search very significantly.

% There are similar transformations for \terminatedAt\ and other forms of \initiatedAt\ rules, and some details that we have omitted for lack of space. The complete code for CER-EC is available at the URL mentioned in the Introduction.

\terminatedAt\ and other forms of \initiatedAt\ rules are handled similarly. We omit these and other details for lack of space. The complete code for LTAR-EC is available upon request. %at \urlBiBTeX{http://www.iit.demokritos.gr/~a.artikis/LTAR.zip}.

\begin{comment}
The domain-independent axioms for \holdsAt\ and \holdsFor\ are such that, for any fluent $F$, \holdsAt$\mathit{(F \val V,T)}$ if and only if time point $T$ belongs to one of the maximal intervals of $I$ such that \holdsFor$\mathit{(F \val V,I)}$. However, the implementation employs various indexing techniques to make the computation of \holdsFor\ more efficient:
briefly, to compute the maximal intervals for which $F\val V$ holds continuously, we find all time-points $T_i$ in which $F\val V$ is initiated, and then, for each $T_i$, we compute the first time-point after $T_i$ in which $F\val V$ is terminated; both steps can make use of the indexing to reduce the search.
Due to space limitations we do not present here the details of the implementation. The complete code for all domain-independent axioms of LTAR-EC is available at \urlBiBTeX{http://www.iit.demokritos.gr/~a.artikis/LTAR.zip}.

We find it convenient to think of short-term activities as durative events occurring over a (not necessarily instantaneous) interval of time points. \happensFor$\mathit{(Act,I)}$ expresses that $I$ is the list of the maximal intervals over which the short-term activity $\mathit{Act}$ occurs;
\happens$\mathit{(Act,T)}$ that $\mathit{Act}$ is occurring at $T$, i.e., that $T$ is a time point belonging to one of the intervals in $I$. Long-term activities are represented as fluents.
\end{comment}

\vspace{-.3cm}
\section{Short-Term Activities}\label{sec:stbr}
\vspace{-.1cm}

Our long-term activity recognition system (LTAR) includes long-term activity definitions in LTAR-EC. The input to LTAR is a symbolic representation of short-term activities. The output of LTAR is a set of recognised long-term activities.
In \cite{artikis09AIAIshort, artikisIJAITshort} we used the first dataset of the CAVIAR project\footnote{{\small \urlBiBTeX{http://groups.inf.ed.ac.uk/vision/CAVIAR/CAVIARDATA1/}}} to perform long-term activity recognition.
This dataset includes 28 surveillance videos of a public space.
The videos are staged --- actors walk around, sit down, meet one another, leave objects behind, fight, and so on.
Each video has been manually annotated in order to provide the ground truth for both short-term and long-term activities.
%The CAVIAR dataset includes the following short-term activities: walking, running, active and inactive. We recognised the following long-term activities: a person leaving an object, a person being immobile, people meeting, moving together, or fighting.

Our preliminary experiments with this dataset, however, showed that the limited dictionary of short-term activity types compromised the recognition of some long-term activities --- it was often impossible to distinguish between certain long-term activities. % --- for instance, the short-term activities of people fighting were often classified as walking or active (due to the absence of an abrupt motion short-term activity), leading, in certain conditions, to the recognition of fighting and meeting, or fighting and moving.
To overcome this problem, in the context of this paper we introduced in the CAVIAR dataset a short-term activity for `abrupt motion': we manually edited the annotation of the CAVIAR videos by changing, when necessary, the label of a short-term activity to `abrupt motion'. This is a form of short-term activity that is recognised by some state-of-the-art recognition systems, such as \cite{kosmo08setnshort}.
A person is said to exhibit an `abrupt motion' activity if he moves abruptly and his position in the global coordinate system does not change significantly --- if it did then the short-term activity would be classified as `running'.
For this set of experiments, therefore, the input to LTAR is:

\begin{compactenum}
 \item[(i)] The short-term activities abrupt motion, walking, running, active (non-abrupt body movement in the same position) and inactive (standing still), together with their time-stamps, ie the video frame in which that short-term activity took place. These activities are mutually exclusive. This type of input is represented by means of the \happens\ predicate --- eg, $\mathit{\happens(abrupt(id_6),}$ $\mathit{ 15560)}$ expresses that $\mathit{id_6}$ moved abruptly at video frame (time-point) $\mathit{15560}$. Short-term activities are represented as events in the Event Calculus in order to use the \initiatedAt\ and \terminatedAt\ predicates for expressing the conditions in which these activities initiate and terminate a long-term activity.
 \item[(ii)] The coordinates of the tracked people and objects as pixel positions at each time-point. The coordinates are represented with the use of the \holdsAt\ predicate --- eg, $\mathit{\holdsAt(coord(id_2)\val (14, 55),\ 10600)}$ expresses that the coordinates of $\mathit{id_2}$ are $\mathit{(14, 55)}$ at time-point (frame number) $\mathit{10600}$.
 \item[(iii)] The first and the last time a person or object is tracked (`appears'/`disappears'). This type of input is represented using the \happens\ predicate. Eg, \linebreak$\mathit{\happens(appear(id_{10}),\ 300)}$ expresses that $\mathit{id_{10}}$ is first tracked at time-point (frame number) $\mathit{300}$. 
\end{compactenum}

Given this input, LTAR recognises the following long-term activities: a person leaving an object, a person being immobile, people meeting, moving together, or fighting. Long-term activities are represented as Event Calculus fluents in order to use the \holdsFor\ predicate for computing the intervals of these activities. Eg, $\mathit{\holdsFor(moving(id_1, id_3)\val\true,}$ $\mathit{ [(0,40), (340,380)])}$ states that $id_1$ was moving together with $id_3$ in the intervals $\mathit{(0,40)}$, and $\mathit{(340,380)}$.

To recognise long-term activities, LTAR processes the input information as follows. First, given input type (i), ie the short-term activities detected at each time-point and recorded using the \happens\ predicate, LTAR computes the maximal duration of each short-term activity, and represents it using the \happensFor\ predicate. Eg, $\mathit{\happensFor(walking(id_5),}$ $\mathit{[(40, 400),}$ $\mathit{(600, 720)])}$ expresses that the maximal intervals for which $\mathit{id_5}$ was walking are $\mathit{(40, 400)}$ and $\mathit{(600, 720)}$. $\mathit{appear}(A)$ and $\mathit{disappear}(A)$ are instantaneous events. (They occur at one time point.)
Second, given input type (ii), LTAR computes the distance between two tracked entities and compares the distance with pre-defined thresholds. Eg, $\mathit{\holdsAt(close(id_3, id_5, 30)\val\true,\ 80)}$ expresses that $\mathit{id_3}$ is `close' to $\mathit{id_5}$ at time $\mathit{80}$ in the sense that their distance is at most $\mathit{30}$ pixel positions. Further, LTAR computes the maximal intervals for which two tracked entities are `close' --- eg, $\mathit{\holdsFor(close(id_3, id_5, 24)\val\true,}$ $\mathit{[(40, 80)])}$  states that  $\mathit{(40, 80)}$ is the maximal interval for which the distance between $\mathit{id_3}$ and  $\mathit{id_5}$ is continuously at most $\mathit{24}$ pixel positions.

Long-term activity recognition is based on a knowledge base of long-term activity definitions. Next we present example definition fragments of LTAR's knowledge base.

%\vspace{-.2cm}
\section{Long-Term Acitivity Definitions}\label{sec:ltbr}

The `leaving an object' activity is defined as follows:
\vspace{-.2cm}
\begin{align}
& \label{eq:leaving-inactive-init}
\begin{mysplit}
\initiatedAt(\mathit{leaving\_object(P,\ Obj)\val\true,\ T}) \leftarrow \\
\quad\happens(\mathit{appear(Obj),\ T}), \\
\quad\happens(\mathit{inactive(Obj),\ T}), \\
\quad\holdsAt(\mathit{close(P,\ Obj,\ 30)\val\true,\ T}), \\
\quad\holdsAt(\mathit{person(P)\val\true,\ T}),\\
\quad\happens(\mathit{appear(P),\ T_0}), \ \mathit{ T_0 < T }
\end{mysplit}\\
& \label{eq:leaving-exit-term}
\begin{mysplit}
\terminates(\mathit{leaving\_object(P,\ Obj)\val\true,\ T} ) \leftarrow \\
\quad\happens(\mathit{disappear(Obj),\ T} )
\end{mysplit}
\end{align}
In the CAVIAR videos an object carried by a person is not tracked --- only the person that carries it is tracked. The object will be tracked, ie `appear', if and only if the person leaves it somewhere. Moreover, objects (as opposed to persons) can exhibit only inactive short-term activity. Accordingly, axiom \eqref{eq:leaving-inactive-init} expresses the conditions in which `leaving an object' is recognised. The fluent recording this activity, \leave$\mathit{(P, Obj)}$, becomes \true\ at time $T$ if $\mathit{Obj}$ `appears' at $T$, its short-term activity at $T$ is `inactive',  there is a person $P$ `close' to $\mathit{Obj}$ at $T$, and $P$ has `appeared' at some time earlier than $T$.
%The $\mathit{appearance}$ fluent records the times in which an object/person `appears' and `disappears'.
Recall that $\mathit{appear(A)}$ is an event that takes place at the first time $A$ is tracked and that the $\mathit{close(A, B, D)}$ fluent is \true\ when the distance between $A$ and $B$ is at most $D$ pixel positions.
%The distance between two tracked objects/people is computed given their coordinates.
The value of $\mathit{30}$ pixel positions was determined from an empirical analysis of CAVIAR.
%
%An object that is picked up by someone is no longer tracked --- it `disappears' --- triggering an $\mathit{exit}$ event, which in turn terminates \leave.

In CAVIAR there is no explicit information that a tracked entity is a person or an inanimate object. Therefore, in the activity definitions we try to deduce whether a tracked entity is a person or an object given, among others, the detected short-term activities. We defined the fluent $\mathit{person(P)}$ to have value \true\ if $P$ has exhibited an active, walking, running or abrupt motion short-term activity since $P$ `appeared'. The value of $\mathit{person(P)}$ is time-dependent because in CAVIAR, the identifier $P$ of a tracked entity that `disappears'  (is no longer tracked) at some point  may be used later to refer to another entity that `appears' (becomes tracked), and that other entity may not necessarily be a person.

Unlike the specification of $\mathit{person}$, it is not clear from the CAVIAR data whether a tracked entity is an object, and for this reason we do not have a fluent explicitly representing that an entity is an object. $\mathit{person(P)\val\false}$ does not necessarily imply that $P$ is an object; it may be that $P$ is not tracked, or that $P$ is an inactive person.
Note finally that axiom~\eqref{eq:leaving-inactive-init} incorporates a (reasonable) simplifying assumption,  that a person entity will never exhibit `inactive' activity at the moment it first `appears' (is tracked). If an entity is `inactive' at the moment it  `appears' it can be assumed to be an object, as in the first two conditions of axiom~\eqref{eq:leaving-inactive-init}. (This assumption is adequate for CAVIAR. Removing it raises further issues we do not have space to discuss fully here.)

The lack of explicit information that a tracked entity is an inanimate object may compromise recognition accuracy in certain conditions. A discussion about the effects of the limitations of CAVIAR's dictionary on recognition accuracy will be presented in the next section.

Axiom \eqref{eq:leaving-exit-term} expresses the conditions in which a \leave\ activity ceases to be recognised.
%(In this paper \linebreak  $\mathit{\terminatedAt( F\val\true,\ T )}$ results in $\mathit{\holdsAt( F\val\false,}$ $\mathit{T{+}1 )}$. In other formalisations we may choose to specify that \linebreak $\mathit{\terminatedAt( F\val\true,\ T )}$ expresses that $F$ has no value at $T{+}1$, ie we are agnostic about the value of $F$.)
In brief, \leave\ is terminated when the object in question is picked up.
An object that is picked up by someone is no longer tracked --- it `disappears' --- terminating \leave.

% immobile
The long-term activity \immobile\ was defined in order to signify that a person is resting in a chair or on the floor, or has fallen on the floor (eg, fainted).
Note that there is no short-term activity in the CAVIAR annotation for the motion of leaning towards the floor or a chair. The absence of such a short-term activity substantially complicates the definition of \immobile, and, as discussed in the next section, sometimes reduces the accuracy of recognising \immobile.
Below is one of the axioms of the \immobile\ definition:
\begin{align}
& \label{eq:immobile-inactive-init}
\begin{mysplit}
\initiatedAt(\mathit{immobile(P)\val\true,\ T}) \leftarrow \\
\quad \mathit{\happensFor(inactive(P),\ Intervals),} \\
\quad  \mathit{(T,\ T_1)\ \in\ Intervals,} \ \mathit{T_1 > T{+}54,}\\
\quad\holdsAt(\mathit{person(P)\val\true,\ T}),\\
\quad \mathit{findall(S,\ shop(S),\ Shops),}\\
\quad\mathit{ \holdsAt(farS( P,\ Shops,\ 24 )\val\true,\ T)}
%\quad \mathit{T_0 < T}
\end{mysplit} %\\
%& \label{eq:immobile-walking-term}
%\begin{mysplit}
%\initiates(\ \mathit{walking(Person),\ immobile(Person)\val \false,\ T}\ )
%\end{mysplit} \\
%& \label{eq:immobile-running-term}
%\begin{mysplit}
%\initiates(\ \mathit{running(Person),\ immobile(Person)\val\false,\ T}\ )
%\end{mysplit} \\
%& \label{eq:immobile-exit-term}
%\begin{mysplit}
%\initiates(\ \mathit{ exit(Person),\ immobile(Person)\val\false,\ T}\ )
%\end{mysplit}
\end{align}
\immobile($P$) is recognised if the following conditions are satisfied. First, $P$ stays inactive for more than $\mathit{54}$ frames (see lines 2--3 of axiom \eqref{eq:immobile-inactive-init}). We chose this number of frames, like all other numerical constraints of the definitions, based on empirical analysis of the CAVIAR dataset. 
Second, $P$ is a person (see line 4 of axiom \eqref{eq:immobile-inactive-init}). 
%ie $P$ has exhibited an active, walking, running or abrupt motion short-term activity since $P$ `appeared'. 
With the use of this constraint we distinguish between an inanimate object, which is inactive since it is first tracked, from an immobile person.
Third, $P$ is not `close' to a shop (see lines 5--6 of axiom \eqref{eq:immobile-inactive-init}). If $P$ were `close' to a shop then he would have to stay inactive much longer than $\mathit{54}$ frames before \immobile\ could be recognised. (Those conditions are specified in other axioms defining \immobile\ not shown here.) In this way we avoid classifying the activity of browsing a shop as \immobile. $\mathit{farS(A, List, D)}$ is \true\ when $A$ is more than $D$ pixel positions away from \emph{every} element of the $\mathit{List}$.

\immobile($P$) is terminated when $P$ starts walking, running or `disappears', ie he is no longer tracked by the video cameras. The relevant axioms for  \terminatedAt\ are straightforward and are not shown here.

% --- see axioms \eqref{eq:immobile-walking-term}--\eqref{eq:immobile-exit-term}:
%
%In a similar way we may express the definitions of other long-term activities.
%It is not difficult to see that the use of the Event Calculus, in combination with the full power of logic programming, allows us to express activity definitions including complex temporal, spatial or other constraints.
%Below we present fragments of the remaining activity definitions of LTAR's knowledge base.
%

% meet

\meet\ (of two persons $P_1$ and $P_2$) is recognised when two people `interact': at least one of them is active or inactive, the other is neither running nor moves abruptly, and the distance between them is at most $\mathit{25}$ pixel positions. In the CAVIAR annotations, this interaction phase can be seen as some form of greeting (eg, a handshake). The rule below shows one set of conditions in which \meet\ is initiated:
\begin{align}
%\label{eq:meet-active-init}
%& \begin{mysplit}
%\initiates(\ \mathit{meeting(P_1,\ P_2)\val\true,\ T}\ ) \leftarrow \\
%\quad\holdsAt(\ \mathit{close(P_1,\ P_2,\ 25)\val\true,\ T}\ ), \\
%\quad \happens(\ \mathit{active(P_1),\ T}\ ),\\
%\quad \holdsAt(\ \mathit{person(P_2),\ T}\ ),\\
%\quad\nbf\ \happens(\ \mathit{running(P_2),\ T}\ ),\\
%\quad\nbf\ \happens(\ \mathit{abrupt(P_2),\ T}\ )
%\end{mysplit}\\
& \label{eq:meet-inactive-init}
\begin{mysplit}
\initiatedAt(\mathit{meeting(P_1,\ P_2)\val\true,\ T})\leftarrow \\
\quad\holdsAt(\mathit{close(P_1,\ P_2,\ 25)\val\true,\ T}),\\
\quad \holdsAt(\mathit{person(P_1),\ T}),\\
\quad \happens(\mathit{inactive(P_1),\ T}),\\
\quad \holdsAt(\mathit{person(P_2),\ T}),\\
\quad\nbf\ \happens(\mathit{running(P_2),\ T}),\\
\quad\nbf\ \happens(\mathit{abrupt(P_2),\ T})
\end{mysplit}
\end{align}
%
%\nbf\ represents negation by failure. 
%\meet\ is initiated when two people `interact': at least one of them is active or inactive, the other is neither running nor moves abruptly, and the distance between them is at most 25 pixel positions. (In axiom~\eqref{eq:meet-active-init} the condition that $P_1$ is `active' at time $T$ implies that $P_1$ identifies a person at time $T$.) In the CAVIAR annotations, this interaction phase can be seen as some form of greeting (eg, a handshake).
\meet\ is terminated when the two people walk away from each other, or one of them starts running, moves abruptly, or `disappears'. The formalisation is straightforward and so omitted here.
 %The axioms representing the termination of \meet\ are similar to axioms \eqref{eq:move-walking-term} and  \eqref{eq:move-running-term}--\eqref{eq:move-abrupt-term}.

The activity \move\ was defined in order to recognise whether two people are walking along together. This activity, like the activities presented so far, could be formalised in terms of \initiatedAt/\terminatedAt\ predicates to specify the conditions in which \move\ starts/ceases to be recognised, and then using the domain-independent axioms of \holdsFor\ to compute the maximal intervals of this activity:
\move\ is initiated when two people are walking and are `close' to each other, and terminated when the people walk away from each other, when they stop moving, ie become active  or inactive, when one of them starts running, moves abruptly, or `disappears'.

A considerably more concise representation of \move, however, can be given directly in terms of \holdsFor:
\begin{align}
& \label{eq:move}
\begin{mysplit}
\holdsFor(\mathit{moving(P_1, P_2)\val\true,\ MovingI }) \leftarrow \\
\quad\holdsFor(\mathit{close(P_1, P_2, 34)\val\true,\ CloseI }),%\ \mathit{CloseI \neq [],} 
\\
\quad \mathit{\happensFor(walking(P_1),\ WalkingI_1),}% \ \mathit{WalkingI_1 \neq [],} 
\\
\quad \mathit{\happensFor(walking(P_2),\ WalkingI_2),}% \ \mathit{WalkingI_2 \neq [],} 
\\
% %\quad\mathit{ (T_1, T_2)\val CloseIntervals \cap WI_1 \cap WI_2 }
% \quad \mathit{\textrm{intersection}(\ WalkingI_1, WalkingI_2,\ WalkingI\ ),}\\
% \quad \mathit{\textrm{intersection}(\ WalkingI,\  CloseI,\ MovingI\ )}
\quad \intersectall\mathit{([WalkingI_1, WalkingI_2, CloseI],\ MovingI)}
\end{mysplit}
\end{align}
$\mathit{CloseI}$ are the maximal intervals in which the distance between $P_1$ and $P_2$ is continuously at most $\mathit{34}$ pixel positions. We compute these intervals using the recorded trajectories of $P_1$ and $P_2$ given as input to LTAR.
\intersectall\ computes the intersection of a list of intervals. 
The implementation of \intersectall\ and other constructs manipulating intervals is available with the source code of LTAR-EC.
According to axiom \eqref{eq:move}, the maximal intervals in which $P_1$ and $P_2$ are \move\ together are produced by the intersection of the intervals in which $P_1$ is `close' to $P_2$, $P_1$ is walking and $P_2$ is walking.

As in the case of \move, we could also have formalised \leave, \immobile\ and \meet\  directly in terms of \holdsFor\ (as opposed to representing these activities in terms of \initiatedAt\ and \terminatedAt\ and then using the domain-independent axioms of \holdsFor\ to compute their maximal intervals). However, formalising \leave, \immobile\ and \meet\ directly in terms of \holdsFor\ is not more concise than formalising these activities  in terms of \initiatedAt\ and \terminatedAt.
For \leave, \immobile\ and \meet\ it is much simpler to identify the conditions in which these activities are initiated and terminated, than identifying all possible conditions in which these activities hold.

%Note that \meet\ may overlap with \move: two people interact and then start \move, ie walk while being `close' to each other. In general, however, there is no fixed relationship between \meet\ and \move.

% fight
%
The last definition of LTAR's knowledge base concerns the \fight\ activity:
\begin{align}
& \label{eq:fighting}
\begin{mysplit}
\holdsFor(\mathit{fighting(P_1, P_2)\val\true,\ FightingI }) \leftarrow \\
\quad \mathit{\happensFor(abrupt(P_1),\ AbruptI),} %\ \mathit{AbruptI \neq [],}
 \\
\quad\holdsFor(\mathit{close(P_1, P_2, 24)\val\true,\ CloseI }),%\ \mathit{CloseI \neq [],}
 \\
\quad \intersectall\mathit{([AbruptI, \ CloseI],\ AbruptCloseI)} \\
\quad \mathit{\happensFor(inactive(P_2),\ InactiveI),}\\
\quad \complement\mathit{(AbruptCloseI,\ InactiveI,\ FightingI) }
%\quad \mathit{\textrm{intersection}(\ AbruptCloseI,\ InactiveI, \ AbruptInactiveCloseI\ ), }\\
%\quad \mathit{\textrm{union}(\ AbruptInactiveCloseI,\ FightingI,\ AbruptCloseI\ ) }
\end{mysplit}
\end{align}
\complement\ is an implementation of the complement operation.
Two people are assumed to be \fight\ if at least one of them is moving abruptly, the other is not inactive, and the distance between them is at most $\mathit{24}$ pixel positions. %We have specified that running initiates \fight\ because, in the CAVIAR dataset, moving abruptly, which is what happens during a fight, is often classified as running.
%\fight\ is terminated when one of the people walks or runs away from the other, or `disappears'.%% --- see axioms \eqref{eq:fight-walking-term}--\eqref{eq:fight-exit-term} below:
%
As in the case of \move, we expressed the definition of \fight\ directly in terms of \holdsFor\ because expressing the conditions in which two people are fighting leads to a more succinct representation than expressing the conditions in which \fight\ is initiated and terminated. %Note that according to axiom~\eqref{eq:fighting}, when a person $P_2$ who is fighting becomes inactive, the period of fighting is terminated, and then initiated again if/when $P_2$ ceases to be inactive.

\vspace{-.4cm}
\section{Experimental Results}
\vspace{-.1cm}

% comment on what constitutes a true positive? Ie does the interval of the recognised activity match exactly that of the ground truth?

We present experimental results on 28 surveillance videos of the CAVIAR project. These videos contain 26419 frames that were manually annotated by the CAVIAR team in order to provide the ground truth for short-term and long-term activities. We edited the original CAVIAR annotation by introducing a short-term activity for abrupt motion. 
%(Samples of the edited annotation of the CAVIAR videos is available with the source code of LTAR.) 
Table \ref{tbl:results} shows the performance of LTAR; it shows, for each long-term activity, the number of True Positives (TP), False Positives (FP) and False Negatives (FN), and the corresponding Recall and Precision values. Long-term activities are recognised with the use of the \holdsFor\ Event Calculus predicate.
%The output of EC and CEC is the same in each case since they are merely different algorithms for computing the same set of maximal intervals.

% RECALL = TP/(TP+FN)
% PRECISION = TP/(TP+FP)

% INCOMPLETE EVENT NARRATIVE

LTAR achieved high Recall and Precision rates, indicating that it may adequately represent complex activities. Perfect Recall and Precision rates were not achieved due to various reasons.
%There are several reasons for not achieving perfect Recall and Precision.
One of these reasons concerns the fact that the narrative of short-term activities (produced by manual annotation, in the present experiments) is incomplete. Eg, the single FN concerning \leave\ is due to the fact that in the video in question the object was left behind a chair and was not tracked. In other words, the left object never `appeared', it never exhibited a short-term activity.

% INCONSISTENT ANNOTATION

Another reason for having FP and FN is the lack of consistency in the annotation of the videos; eg, the long-term activity of people walking in the same direction while being `close' to each other is not always classified as \move\ (this type of inconsistency leads to FP concerning the recognition of \move), the short-term activity of people being active is sometimes classified as walking (eg, leading to FN in the recognition of \meet), and so on. %(An evaluation of the manual annotation of the CAVIAR dataset is presented in \cite{list05short}.)

% LACK OF DETAIL

The most important reason for not achieving perfect Recall and Precision in the CAVIAR dataset concerns the limited dictionary of short-term activities and context variables with which the tracked activity is represented.
The recognition of \immobile, for instance, would be much more accurate if there were a short-term activity for the motion of leaning towards the floor or a chair. In the absence of such an activity, the recognition of \immobile\ is primarily based on how long a person is inactive.
In the CAVIAR videos  a person who falls on the floor or rests in a chair stays inactive for at least $\mathit{54}$ frames. Consequently LTAR recognises \immobile\ if, among other things, a person stays inactive for at least $\mathit{54}$ frames. There are situations, however, in which a person stays inactive for more than $\mathit{54}$ frames and has not fallen on the floor or sat in a chair: people watching a fight, or just staying inactive waiting for someone. It is in those situations that we have the FP concerning \immobile.

For similar reasons we did not achieve perfect Recall and Precision in the recognition of \meet; it is impossible to  define this activity precisely due to the absence of a short-term activity for `greeting'.

\begin{table}[t]
\caption{Experimental Results.}\label{tbl:results}%\vspace{-.2cm}
\begin{center}
\renewcommand{\arraystretch}{0.9}
\setlength\tabcolsep{3pt}
\begin{tabular}{lccccc}
\hline\noalign{\smallskip}
\multicolumn{1}{c}{\textbf{Behaviour}} & \multicolumn{1}{c}{\textbf{TP}} & \multicolumn{1}{c}{\textbf{FP}} & \multicolumn{1}{c}{\textbf{FN}} & \multicolumn{1}{c}{\textbf{Recall}} & \multicolumn{1}{c}{\textbf{Precision}}  \\
%\noalign{\smallskip}
\hline
\noalign{\smallskip}
leaving object  & 4 & 0 & 1 & 0.8 & 1 \\[3pt]       % THE SAME AS AIAI PAPER
immobile    & 9 & 8 & 0 & 1 & 0.52 \\[3pt]      % THE SAME AS AIAI PAPER
meeting     & 6 & 1 & 3 & 0.66 & 0.85 \\[3pt]
moving      & 15 & 3 & 2 & 0.88 & 0.83 \\[3pt]  % FP ARE VIDEOS 21, 22 AND 27-2. FN ARE 12 AND 19
fighting    & 6 & 0 & 0 & 1 & 1  \\
\hline\vspace{-1.3cm}
\end{tabular}
\end{center}
\end{table}

A particular refinement of CAVIAR's dictionary --- the introduction of a short-term activity for abrupt motion --- considerably increased LTAR's recognition accuracy.
More precisely, compared to our earlier results \cite{artikis09AIAIshort,artikisIJAITshort}, the introduction of abrupt motion reduced the number of FP regarding \move\ and \meet.
In the original annotation of CAVIAR, the short-term activities of people \fight\ were sometimes classified as walking or active. In the first case LTAR incorrectly recognised \move, because two people were walking while being `close' to each other, while in the second case LTAR incorrectly recognised \meet\ (in addition to recognising \fight), because two people were active while being `close' to each other. Labelling the short-term activities of people \fight\ as abrupt motion resolved this issue, because abrupt motion does not initiate \move\ or \meet.

In addition to increasing the recognition accuracy of \move\ and \meet, the introduction of abrupt motion eliminated FP and FN regarding \fight.
Moreover, the introduction of abrupt motion did not increase FP or FN in the recognition of the other long-term activities.

%Note that the introduction of abrupt motion could, in some cases, produce FP concerning \fight. This would happen if a person moved abruptly (say fainted and fell on the floor) while being `close' to another person that was not inactive at the time. In this case LTAR would incorrectly recognise \fight. Such a combination of short-term activities does not take place in the CAVIAR dataset.

Similar to introducing abrupt motion, we could have enhanced CAVIAR's dictionary by including activities for greeting a person, falling on the floor, etc, and variables explicitly representing that a tracked entity is an object.
We did not do this because we are not aware of any short-term activity recognition systems that detect such activities and explicitly represent the aforementioned type of information. In contrast, there are systems that detect abrupt motion --- eg, see \cite{kosmo08setnshort}. We expect that a finer classification of short-term activities and the addition of context variables such as the one mentioned above, will, under certain circumstances, increase the overall activity recognition accuracy, provided that the long-term activity definitions are updated accordingly.

We should like to point out that the issues identified above do not always compromise recognition accuracy. Eg, the lack of explicit information that a tracked entity is an object did not the affect the recognition accuracy of \leave\ in the 28 CAVIAR videos. This lack of information would have led to FP in the recognition of \leave\ in certain conditions, but these conditions did not arise in the CAVIAR videos.
Similarly, the lack of consistency in the annotation of activities, and the incompleteness of short-term activity narratives do not always lead to FP or FN.
In any case, in the next section we discuss ways of addressing these issues.

Concerning recognition efficiency, we were able to recognise each long-term activity in less than 1 second CPU time, given as input around 1800 temporally sorted short-term activities representing, on average, a CAVIAR video, on an Intel Core i7 920@2.67GHz with 6 GB RAM running Linux Kernel 2.6. Ways to further improve recognition efficiency are presented in the following section.

\vspace{-.4cm}
\section{Discussion}
\vspace{-.1cm}

Numerous recognition systems have been proposed in the literature. In this section we focus on long-term activity (high-level) recognition systems that, similar to our approach, exhibit a formal, declarative semantics.

A well-known system for activity recognition is the Chronicle Recognition System (CRS)\footnote{\urlBiBTeX{http://crs.elibel.tm.fr/}}. A `chronicle' can be seen as a long-term activity --- it is expressed in terms of a set of events (short-term activities in our example), linked together by time constraints, and, possibly, a set of context constraints.
The language of CRS relies on a reified temporal logic, where propositional terms are related to time-points or other propositional terms. Time is considered as a linearly ordered discrete set of instants. The language includes predicates for persistence and event absence.
Details about CRS may be found on the web page of the system and \cite{dousson07short}.
%  (Ghallab 1996\cite{ghallab96});(Dousson 2002\cite{dousson02});(Dousson and Maigat 2007\cite{dousson07}).

The CRS language does not allow mathematical operators in the constraints of atemporal variables. Consequently, the computation of the distance between two people/objects cannot be computed. 
CRS, therefore, cannot be directly used for activity recognition in video surveillance applications. More generally, CRS cannot be directly used for activity recognition in applications requiring any form of spatial reasoning, or any other type of atemporal reasoning. These limitations could be overcome by developing a separate tool for atemporal reasoning that would be used by CRS whenever this form of reasoning was required. To the best of our knowledge, such extensions of CRS are not available. Clearly, the computational efficiency of CRS, which is one of the main advantages of using this system for activity recognition, would be compromised by the integration of an atemporal reasoner.

Hakeem and Shah \cite{shah07AIJ} have presented a hierarchical event
representation for analysing videos. The temporal relations between the
sub-events of an event definition (or activity, in the terminology of
this paper) are represented using the interval algebra of \cite{allen94} and an extended form of the CASE representation \cite{fillmore68short} originally used for the syntactic analysis of natural languages.

In our approach to activity recognition, the availability of the full power of logic programming is one of the main attractions of employing the Event Calculus as the temporal formalism. It allows activity definitions to include not only complex temporal constraints --- LTAR-EC is at least as expressive as the CRS language and the extended CASE representation with respect to temporal representation --- but also complex atemporal constraints.
Moreover, when necessary more expressive Event Calculus dialects may be adopted (see, eg, \cite{miller00ashort}).

Shet et al.~have presented a logic programming approach to activity recognition. See \cite{davis05AVSSshort,davis07CVPRshort} for two recent publications. These researchers have presented activity definitions concerning theft, entry violation, unattended packages, and so on. A distinguishing feature of our approach with respect to this line of work concerns the fact that we use the Event Calculus for temporal representation and reasoning.
The temporal aspects of the definitions of Shet, Davis et al.~are crudely represented ---
eg, there are no rules for computing the intervals in which a long-term activity takes place. In contrast, the Event Calculus has built-in axioms for complex temporal representation, including the formalisation of inertia, durative events, events with delayed effects, etc, which help considerably the system designer develop activity definitions.
Shet and colleagues stated that ``[i]n the future we would like to extend this system to reason explicitly about temporal information thus helping us [..] to define models for and recognise human activities within a single framework'' \cite[p.8]{davis07CVPRshort}.
To the best of our knowledge, they have not developed a system for explicit temporal representation and reasoning since.

Shet and colleagues have incorporated in their logic programming framework a mechanism for reasoning over rules and facts that have an uncertainty value attached. We aim to extend our work by allowing for uncertainty values in the rules of activity definitions in order to address, to a certain extent, the issues arising from incomplete short-term activity narratives, inconsistent annotation of short-term and long-term activities, and a limited dictionary of short-term activities and context variables.

Paschke and colleagues \cite{paschke08} have also proposed the use of an Event Calculus dialect for event recognition. This dialect and LTAR-EC have numerous differences. For example, unlike LTAR-EC, there is no support in the dialect of Paschke et al for multi-valued fluents --- only Boolean fluents are considered.
Moreover, the treatment of intervals is quite different. The Event Calculus dialect of Paschke and colleagues, for instance, does not include axioms for recognising an `on-going' long-term activity, ie a activity that started taking place at some earlier time-point and still holds. 
%We see this as a serious limitation. 
There are also very significant differences in the implementations.

Apart from the numerous differences in expressiveness and implementation, a key contribution of the work presented here, as we see it,  is that we have illustrated the expressiveness of the Event Calculus for complex activity recognition on a benchmark example, showed a range of different types of definition, and evaluated the adequacy of our representation empirically.
We expect that the example itself will be a valuable resource in future uses of the Event Calculus for activity/event recognition.

%All issues identified in the previous section --- incomplete short-term activity narratives, lack of consistency in the annotation of activities, limited dictionary of short-term activities --- compromise recognition accuracy in all reviewed recognition systems. A logic programming implementation of the Event Calculus, however, may cope more effectively with the latter issue due to the increased available expressiveness. For example, LTAR-EC allows for a more precise approximation of \immobile\ than CRS.

A logic programming approach to activity recognition has, among others, the advantage that machine learning techniques can be directly employed for developing/refining  activity definitions. An area of current work is the use of abductive and inductive logic programming techniques 
%as proposed, eg, in \cite{ray09}, 
for learning activity definitions. Details about this line of work are given in \cite{artikisIJAITshort}.

% =======

LTAR-EC does not currently store the outcome of query computation, ie the intervals of the recognised activities. Consequently, LTAR-EC often performs unnecessary computations, re-computing activity intervals that it already computed but did not store.
We are currently experimenting to find the most effective options for caching in LTAR-EC, including those presented in \cite{chittaro96}.

\vspace{-.3cm}

\section*{Acknowledgements}
\vspace{-.1cm}

This work was supported partly by the EU PRONTO Project (FP7-ICT 231738).

\vspace{-.3cm}

\bibliographystyle{plain}
%\bibliography{../../../aabib,temp}

\end{document}